# Health Analytics: a systematic review of approaches to detect phenotype cohorts using electronic health records

Norman Hiob, Stefan Lessmann

School of Business and Economics, Humboldt-University of Berlin,

Unter-den-Linden 6, 10099 Berlin, Germany

[normanhiob@me.com, stefan.lessmann@hu-berlin.de]

**Abstract**

The paper presents a systematic review of state-of-the-art approaches to identify patient cohorts using electronic health records. It gives a comprehensive overview of the most commonly detected phenotypes and its underlying data sets. Special attention is given to preprocessing of input data and the different modeling approaches. The literature review confirms natural language processing to be a promising approach for electronic phenotyping. However, accessibility and lack of natural language process standards for medical texts remain a challenge. Future research should develop such standards and further investigate which machine learning approaches are best suited to which type of medical data.

# 1 Introduction

The advent of Electronic Health Records (EHR) has led to an increasing amount of valuable data. In the USA alone, adoption of EHR in private acute care hospitals increased from 9 percent in 2008 to 59 percent in 2013 (Charles et al. 2014). Those data can be used for early mortality prediction, identification of risk factors, drug testing, surveillance, clinical trial recruitment, outcome prediction and disease detection. Traditional approaches rely on manual chart review which is time consuming, expensive and error-prone. Furthermore, the process of extracting meaningful information is often complex and time consuming since information are hidden in multiple data sources. Machine learning (ML) algorithms might offer new opportunities to address this issue. Especially rich textual resources have the potential to provide great value for electronic phenotyping. ML algorithms allow for a quick scan of large volumes of data with minimal human effort at relatively low cost. Furthermore, scalability leaves more time for humans do other things and thus increases productivity. It also allows for near-real time monitoring and quick intervention. Therefore, this paper aims to answer the following three research question in an extensive way:

- What data generate the best predictive outcome?
- Which machine learning approaches achieve the best results?
- What are potential gaps for further research?

The paper is sectioned in five parts. Section one explains the methodology of this paper and gives a brief overview of the phenotypes of interest. Following, section two will examine data sources, data structure and data selection for both structured and unstructured data. Section three will then describe the preprocessing of data with special emphasis on data cleaning, data transformation and natural language processing. Section four outlines modeling approaches currently used for electronic phenotyping. Section five will answer the research questions and give an outlook into further research.

## 2 Method

### 2.1 Design

This paper reviews existing literature and builds on the state-of-the-art review of Shivade et al. from 2014 by examining the years 2015-2017. The literature review followed a three step-process. First, 44 articles were selected on title and abstract review (24 of JBI and 20 of JAMIA). Next, title and abstract of relevant references were reviewed and a total of 13 articles were added. Finally, duplicates were removed and the full text of the remaining articles was read, resulting in a total of 23 articles. Figure 1 shows the search flow chart for study inclusion. All publications in (1) ‚Journal of Biomedical Informatics' (JBI) and in (2) ‚Journal of American Medical Informatics Association' (JAMIA) were manually reviewed from the years 2015-2017. In contrast with Shivade et al. (2014) articles in ‚Proceedings of the Annual American Medical Informatics Association Symposium' and in ‚Proceedings of the AMIA Clinical Research Informatics Conference' were not considered due to the limited time and scope of this paper. In contrast to Shivade et al. (2014) only references from the years 2015-2017 were considered, due to the limited time and scope of this paper.

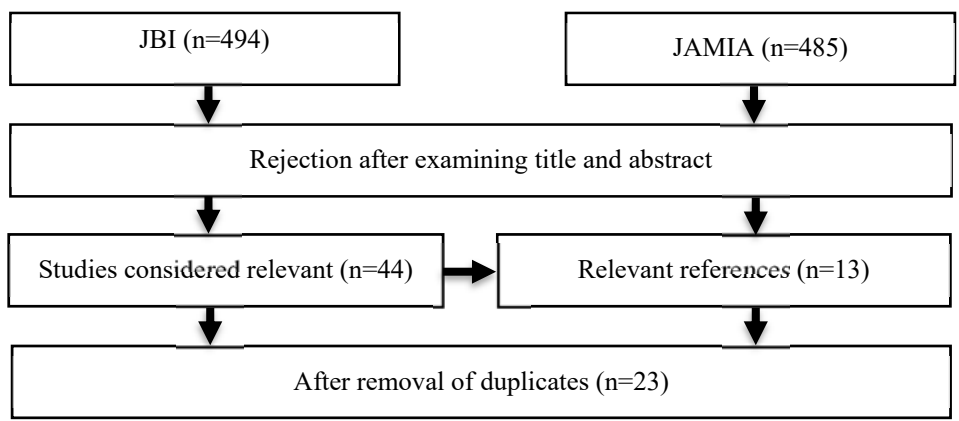

**Figure 1: Search flow chart for study inclusion**

Studies were included which (1) described the identification of patient cohorts with a particular medical condition or disease, (2) characterized solutions for clinical trial recruitment, (3) compared phenotype approaches, described different techniques, or explored variations of data sources to detect phenotype cohorts. Studies were discarded, when they (1) did not identify patient cohorts with a medical condition or (2) did not use EHR data.

## 2.2 Phenotypes

The definition of a 'phenotype' varies across literature and mostly depends on the area of research. In biology, the term often refers to the observable traits of an organism, such as morphology, physiology and behavior (Hancock 2014). Phenotypes are the result of the expression on an individual's genetic code, called 'genotype', which is a gene or a set of genes that are influenced in response to environmental factors. However, gene-association is a different area of research and not the primary focus of this paper. This paper refers to the term 'phenotype' as mostly used in a medical context: to describe a patient, or cohort of patients, with an observable medical condition or disease.

| Phenotype | Studies [1] | Sources |
|---|---|---|
| Diabetes | 7 | Agarwal et al. (2016), Anderson et al. (2015), Spratt et al. (2016), Wei et al. (2015), Yahi and Tatonetti (2015), Zeng et al. (2017), Zhong et al. (2016) |
| Cancer | 7 | Kasthurirathne et al. (2016), Kinar et al. (2016), Kocbek et al. (2015), Kocbek et al. (2016), Li et al. (2015), Osborne et al. (2016), Wei et al. (2015) |
| Heart Failure | 4 | Agarwal et al. (2016), Choi et al. (2016), Evans et al. (2016), Wang et al. (2015) |
| Atrial Fibrillation | 3 | Wang et al. (2016), Wei et al. (2015), Yahi and Tatonetti (2015) |
| Arthritis | 2 | Wei et al. (2015), Yu et al. (2015) |
| Thrombosis [2] | 2 | Kocbek et al. (2016), Rochefort et al. (2015) |
| Others [3] | 7 | Ichikawa et al. (2016), Kocbek et al. (2016), Martinez et al. (2015), Padula et al. (2016), Teixeira et al. (2016), Wei et al. (2015), Yu et al. (2015) |

**Table 1 Top six phenotypes**

Extensive research has focused on classifying patient with Type1 and Type 2 diabetes (Zeng et al. 2017, Spratt et al. 2016, Zhong et al. 2016, Anderson et al. 2015) while other studies focused on cancer (Kasthurirathne et al. 2016, Kinar et al. 2016, Osborne et al. 2016, Kocbek et al. 2015, Li et al. 2015). Although cancer is mostly used as a generic term there were 5 different types of cancers described in a total of four studies, which include lung cancer, breast cancer, gastric cancer, colon cancer and colorectal cancer. In contrast to Shivade et al. (2014), Table 1 shows the top six phenotypes only. Due to the limited number of studies, only phenotypes mentioned at least twice were listed individually. The most commonly classified phenotypes include heart failure,

---

[1] Duplicates possible, as some studies described novel approaches for more than one class (phenotype)

[2] Thrombosis includes deep vein thrombosis and pulmonary embolism

[3] Others include: hyperuricemia, pneumonia, fungal, decubitus, hypertension, Alzheimer's, gout, HIV, multiple sclerosis, Parkinson and coronary artery disease

atrial fibrillation, arthritis and thrombosis. Phenotypes that were mentioned only once are summarized in table 2.1 as 'others'. Some studied described generic approaches which can be used to detect several diseases (Agarwal et al. 2016, Kocbek et al. 2016, Wei et al. 2015). For example, Wei et al. (2015) developed a multiclass classifier that detects atrial fibrillation, Alzheimer, cancer, gout, HIV, multiple sclerosis, Parkinson, arthritis and diabetes.

# 3 Data Selection

In order to outline phenotype modeling approaches comprehensively, structure and characteristics of the data will be outlined briefly. This section will show data usage has shifted from single data sources to multiple data sources. However, accessing EHR data may be difficult due to laws governing data security and data protection.

More than 65 percent of the studies obtained data from the USA while others received data from Australia, China, Canada, Israel, the UK and Japan (see Appendix). Table 2 gives a comprehensive overview of the different data sources used to obtain medical data. More than 50 percent of the studies obtained EHR from an academic medical center which is usually tied to a specific university. Health care systems, which refer to a network of hospitals, clinics, care centers and physicians, served in 30 percent of the cases as a source for EHR. Only 9 percent used data from governmental agencies. Other data sources include health insurance companies (Ichikawa et al. 2016) and publicly available online sources (Yu et al. 2015). Although online knowledge (e.g. Wikipedia, Medscape) provide substantial information, it is unknown whether the online knowledge source may be sufficient for some phenotypes (Yu et al. 2015). The time frame also varies from one year, over four years, to up to ten years (see Appendix). Size of data set is expressed in terms of incidents (e.g. encounters, visits, hospitalizations, admissions) and patients. Some studies refer to patients as person, children or individuals, ranging from 480 to 2.3 million. The diversity of population, in terms of origin, timeframe and sizes, poses an underlying challenge since model performance is not directly comparable.

| Data source | Studies | F(x) | Sources |
|---|---|---|---|
| Academic Medical Center | 12 | 0.52 | Agarwal et al. (2016), Choi et al. (2016), Li et al. (2015), Martinez et al. (2015), Padula et al. (2016), Rochefort et al. (2015), Teixeira et al. (2016), Wang et al. (2016), Wei et al. (2015), Zhong et al. (2016), Spratt et al. (2016), Yahi and Tatonetti (2015) |
| Health Care System | 7 | 0.30 | Anderson et al. (2015), Evans et al. (2016), Kinar et al. (2016), Kocbek et al. (2016), Kocbek et al. (2015), Kasthurirathne et al. (2016), Wang et al. (2015) |
| Government Agency | 2 | 0.09 | Osborne et al. (2016), Zeng et al. (2017) |
| Others | 2 | 0.09 | Yu et al. (2015), Ichikawa et al. (2016) |

**Table 2 Comprehensive overview of EHR data sources**

Previous research has made use of single data sources (Hripcsak et al. 2002, Nguyen et al. 2010) predominantly focusing on ICD-9 codes, which contain standardized information about patient diagnosis and procedures. More recently data linkage, linkage of multiple data sources to an individual patient ID, has emerged (Sorace et al. 2012) which has shown to improve classification performance (Teixeira et al. 2016, Kocbek et al. 2015). Table 3 gives a comprehensive overview of the number of studies using multiple data sources across the top six phenotype classes.

| Data<br>Class (Phenotype) | structured | | | | | | | unstructured | | | Σ | % |
|---|---|---|---|---|---|---|---|---|---|---|---|---|
| | Demographics | Clinical | Lab tests | Medications | Diagnosis | Treatment | Vitals | Notes | Images | Others | | |
| Diabetes | 4 | 4 | 6 | 6 | 4 | 1 | - | 5 | - | 1 | 31 | **0.27** |
| Cancer | 4 | 2 | 1 | 1 | - | - | - | 5 | 1 | - | 16 | **0.14** |
| Heart failure | 1 | 3 | 3 | 3 | 4 | 2 | - | 3 | - | - | 19 | **0.16** |
| Atrial fibrillation | 1 | 2 | 2 | 2 | - | - | 1 | 2 | - | - | 10 | **0.09** |
| Arthritis | 2 | 2 | - | - | - | - | - | 2 | - | - | 4 | **0.03** |
| Thrombosis | 2 | 2 | - | - | 1 | - | - | 2 | - | - | 6 | **0.05** |
| Others | 3 | 8 | 1 | 7 | 1 | - | 1 | 9 | - | - | 30 | **0.26** |
| Σ | 17 | 23 | 13 | 19 | 10 | 3 | 2 | 28 | 1 | 1 | 116 | 1.0 |
| % | **0.13** | **0.21** | **0.11** | **0.17** | **0.08** | **0.03** | **0.02** | **0.24** | **0.01** | **0.01** | 1.0 | |
| | | | | | 0.74 | | | | 0.25 | 0.01 | | |

**Table 3 Number of sources used across the top six phenotypes**

*Structured data*, which includes demographics, clinical data, laboratory tests, diagnosis, treatments and vital signs, accounts for 74 percent of all data. Demographics hereby refer to basic attributes such as gender, age, country, language, religion and ethnic origin. Clinical data, which includes administrative data (e.g. financial data and ICD-9 codes), are by far the most commonly used form of structured data.

However, ICD-9 codes lack specificity in the information transmitted in the code. For example, if a patient is seen for treatment of a burn on the right leg, the ICD-9 code does not distinguish that the burn is on the right leg. If the same patient is seen for another burn on the left leg the same ICD-9 code would be used. Misclassification of ICD-9 codes poses another problem, e.g. if the prescribed medication is not for the patient itself, but for someone else (e.g. a friend or family member). This is in line with previous studies which have shown that the use of ICD-9 codes is not sufficient to identify phenotype cohorts (Birman-Deych et al. 2005) as they often lead to missing patients (Evans et al. 2016)

*Unstructured data* account for about 25 percent of all data. However, looking at all data sources on an individual level shows that notes are the most commonly used form of data. Notes include problem lists, admission notes, progress reports, consulting notes, discharge summaries, and history of examination. Structured data are easy to read and use, but the problem is these codes are often stored long after patients have been discharged. Notes could help to deliver more up-to-date information on the patient. However, unstructured data create a new challenge which is due to their heterogeneity, missing values and different standardization. Much of this information is hidden in free text which often requires additional tools (e.g. NLP for extracting notes) to bring them into computer readable format (Wie & Denny 2015). Text mining poses a promising strategy for building ML classifiers that benefit of the richness of such records. For example, adding radiology questions and reports to ICD-9 codes has been shown to significantly improve model performance (Kocbek et al. 2015).

Mining large datasets, consisting of multiple data sources, presents a number of challenges. It is expensive, time consuming and poses the risk of high dimensionality. Several studies addressed this issue by applying a sample method to the original dataset. Sampling is a common technique to reduce dimensionality. It aims to select a subset of data which allows for generalization from the sample to the entire population within a computable margin of error. Hence, sample size and other characteristics of the sample determine whether the sample is a good representation of the dataset. Random sampling without replacement is a common sampling technique applied by a number of studies (Zeng et al. 2017, Agarwal et al. 2016, Wang et al. 2016, Kocbek et al. 2015). It describes the selection of *n* distinct features from *N* samples in the dataset without replacement, thus a sample may not occur twice. The simplicity of the algorithm and exclusion of biases in the selection process are the two main advantages of this approach. Random sampling with replacement, in which a sample is selected by unit with equal probability of selection, was only applied by Teixeira et al (2016). Sometimes imbalanced data could pose a challenge to machine learning algorithms, which usually perform better over balanced data. Ichikawa et al. (2016) addressed the issue of imbalanced data by applying random undersampling which has proven to outperform other methods of handling imbalances medical data (Drummond & Holte 2003). Stratified sampling is a less common type of sampling which also addresses the issue of imbalanced data. Spratt et al. (2016) subdivided the dataset into $n = 4$ strata where sampling was performed individually on each subset (strata). The combination of samples from all four strata were combined into a final subset for further analysis, thus helped to produce robust estimates.

Next the dataset, or the subset of the original dataset, needs to be prepared for the application of ML algorithms. Supervised learning requires two types of datasets: (1) a training set, which requires some labeled training data, and (2) test set, which is used to apply the predictive model. The easiest way to generate a test and training set is to randomly split the data set into two parts (most studies used 2/3 as rule of thumb).

Test set refers to the data set that is used to apply the predictive model. Hence, it serves as a reference against which the classification model is evaluated. Training set, also referred to as ‚gold-standard', is a labeled dataset where the classification is made known to the system. Most studies applied an expert algorithm, which describes manual chart reviews by one or more experts (Agarwal et al. 2016, Evans et al. 2016, Kocbek et al. 2016). The most common procedure is to have two to three domain experts who sample patients into categories[4] describing the presence or absence of the phenotype. The decision is based on expert knowledge considering phenotypic criteria such as disease related medications, diagnosis or laboratory tests (Zeng et al. 2017). Inter-annotator agreement was measured by Cohen's Kappa statistics, which 'compares the results of the first annotator to the results of the other annotators' (Wang et al. 2016, Martinez et al. 2015, Wang et al. 2015). ICD-9 codes, which convey information about patient diagnosis and procedures, were used in some studies to select training and test data (Padula et al. 2016, Wang et al. 2015). Kocbek et al. (2015) used International Classification of Disease, Tenth-Division (ICD-10) codes as ground truth, which is an extension of ICD-9 codes and therefore contains more detailed

---

[4] Usually patients are categorized into positive, negative and unconfirmed cases

information. Yet, high missing rates remain a problem. Only one study used a labeled cohort from the eMERGE network as ground truth (Yahi & Tatonetti 2015).

## 4 Preprocessing

Preprocessing describes the process of preparing input data for the application of ML algorithms. This process typically involves (1) data cleaning, (2) data transformation and (3) data reduction. Activities in the process are interrelated, thus data preprocessing does not necessarily follow a linear process. Furthermore, decisions in a later stage may also affect the suitability of preprocessing activities in an earlier stage. Data cleaning and data transformation will be discussed for structured and unstructured data separately since they pose substantial differences.

### *4.1 Structured Data*

#### 4.1.1 Data Cleaning

EHR contain highly personal information which are protected by law. Although medical data protection laws differ (e.g. between the EU and the USA) actions need to be taken to protect patient privacy. This includes the anonymization of patient data (Kinar et al. 2016, Teixeira et al. 2016, Wie et al. 2015) e.g. by removing visit dates (Anderson et al. 2015). However, most studies used already de-identified data sets which did not require any additional actions for anonymization. Another common pre-step is the aggregation of data. That includes combining multiple visits into one data point representing each patient (Anderson et al. 2015) and collapsing multiple readings (e.g. blood pressure) from the same day to their median value (Teixeira et al. 2016). Only Ichikawa et al. (2016) addressed the issue of missing data by applying knnImpute with $k = 5$, which is an input method that utilizes the k-nearest neighbor (KNN) method. knnImpute replaces missing values in data with the corresponding weighted mean of the $k$ nearest neighbors. Interestingly, removal of outliers was not mentioned in any of the reviewed papers. This might be because outliers (e.g. abnormal lab tests) are often used as a proxy for identifying a certain medical condition (Agarwal et al. 2016).

#### 4.1.2 Data Transformation

Data transformation describes the process of converting data into an appropriate format for the application of machine learning algorithms. Variability of the underlying data can influence the results of some data mining algorithms. Normalizing numeric variables can help to standardize the effect each variable has on the result. Especially algorithms that rely on distance measure, such as Neural Nets, Support Vector Machines and k-nearest neighbor, benefit from standardization.

Ichikawa et al. (2016) proposed a z-score transformation to put continuous variables from different sources onto the same scale. The standard z-score is

$$Z = \frac{X-\mu}{\sigma} \qquad (1)$$

where $\mu$ is the mean of the population and $\sigma$ is the standard deviation of the population. Data smoothing is another technique commonly used in data preparation, which typically involves discretization of data. If a dataset consists of huge a number of continues variables, model building for such data can be challenging or extremely inefficient. Discretization categorizes data into bins to capture characteristics that could be helpful to removes noise and handle inconsistencies. Furthermore, many data mining algorithms (e.g. rule-based algorithms, decision tree) operate with discrete attributes only. Several studies applied this technique by abstracting dates to months, and time to a.m./p.m. (Kocbek et al. 2016). Martinez et al. (2015) followed a similar approach by normalizing dates into a 'DATE' feature and numbers into 'NUMBER' feature. Some studies proposed to transform laboratory test results into categorical variables (Agarwal et al. 2016, Yahi & Tatonetti 2015). However, discretization poses two underlying challenges: how to choose the number of intervals and how to choose the width of each interval. Yahi & Tatonetti (2015) addressed this issue by applying statistical enrichment analysis which associates each laboratory test with terms from a given ontology – e.g. 'blood glucose low' or ,blood glucose high'. Overall preprocessing of structured data has only played a minor role. However, preprocessing of unstructured data is explained in more detail in the following section.

### *4.2 Unstructured Data*

The challenge with unstructured data is that there is no defined variable. Thus, unstructured data need to be transformed into a structured format. Narrative text, such as problem lists, admission notes, discharge summaries, progress reports and history of physical examination, was the most commonly used type of unstructured data. Images were used in one study only. Hence, this section will focus on Natural Language Processing (NLP) primarily.

#### 4.2.1 Ontology-Based Approaches

Processing clinical text is rather complex for several reasons: (1) clinical text is often short and not using the correct rules of grammar, (2) texts are often dictated (e.g. discharge summaries), (3) progress notes often contain slang, acronyms and abbreviations, (4) text may contain misspellings especially when spell check is not available, (5) presence of alphabets other than English (e.g. prescriptions) and (6) text is intentionally structured in tables (Liu et al. 2001).

In order to extract information from clinical narrative, text needs to be transformed into a computer-understandable format. Hence, mining free text requires some form of coding which is done by mapping free-text to an ontology such as the BioPortal from the National Center for Biomedical Ontology (Agarwal et al. 2016), the Metathesaurus of the Unified Medical Language System (UMLS) from the US National Library of Medicine (Kocbek et al. 2016, Kocbek et al. 2015, Martinez et al. 2015), the Facility Oncology Registry Data Standards (Osborne et al. 2016), or equivalent. cTAKES (Lin. et al. 2015), KnowledgeMap (Denny et al. 2014) and MedEx (Teixeira et al. 2016, Wie et al. 2015) are common tools to extract concepts from clinical text. However, MetaMap (Agarwal et al. 2016, Osborne et al. 2016) by far is the most commonly used NLP system to extract medical concepts. Table 4 shows NLP systems used in recent research and their purpose of contents.

| Clinical NLP System | Purpose |
|---|---|
| MetaMap | Highly configurable program to map biomedical text to UMLS Metathesaurus concepts |
| cTAKES | Major clinical Text Analysis and knowledge Extraction Map |
| KnowledgeMap | Content management system to enhance the delivery of medical education contents |
| MedEx | Free-text clinical records to recognize medication names and signature information |

**Table 4 Clinical NLP systems (Aggarwal & Reddy 2015, p. 227)**

Teixeira et al. (2016) used KnowledgeMap to extract non-negated, disease related UMLS concepts from sections, which were previously identified using SecTag (Denny et al. 2014). SecTag is an algorithm which identifies note section headers. Often clinical notes contain substructures that separate them into different segments, such as 'Family Medical History' or ‚History of Present Illness'. Thus, SecTag helps to reduce false positives.

MedEx can be used to extract medications from clinical text (Teixeira et al. 2016, Wie et al. 2015). Medications are one of the most important clinical data in EHR, often recorded in free-text. Therefore, they are not accessible to further applications which rely on coded data. MedEx solves this problem by extracting medications from clinical narratives. Disease related medications can be determined by using the Medication-Indication resource-High Performance Subset (MEDI-HPS), which comprises indications of medications from RxNorm, SIDER 2, MedlinePlus, and Wikipedia into a single resource (Teixeira et al. 2016).

To extract meaningful information it is necessary to understand the relation between the extracted concepts (such as diagnosis, drugs and symptoms). MetaMap addresses this issue by obtaining semantic information and reducing the number of synonyms through disambiguation. It uses a 'knowledge-intensive approach based on symbolic, NLP and computational-linguistic techniques' (Martinez et al. 2015). Figure 2 shows a sample of a MetaMap-annotated phrase.

> „*replaced with a right frontal approach*".
>     Meta Mapping (701):
>         748 C0559956: Replaced (Replacement) [Functional Concept]
>         748 C0205090: Right [Spatial Concept]
>         778 C2316681: Frontal approach [Functional Concept]

**Figure 2 Sample of MetaMap-annotated phrase (Kocbek et al. 2016)**

Negation is an important concept in extracting valuable knowledge from free text. Some NLP systems have already built in modules for negation analysis such as the NegEx algorithm, which is a part of MetaMap. NegEx is the most commonly applied module to recognize the presence or absence of negations (e.g. no elevated blood pressure) in clinical text (Kocbek et al. 2015, Martinez et al. 2015). It uses regular expressions to filter out sentences that falsely appear to be negations. Hence, it helps to prevent to misclassify phenotype cohorts. Figure 3 shows an example of a NegEx-annotated phrase.

> ''No growth detected" would be as follows:
>     NEGATIONS:
>         Negation Type: nega
>         Negation Trigger: no
>         Negation PosInfo: 337/2
>         Negated Concept: C0220844:growth
>         Concept PosInfo: 340/6

**Figure 3 Sample of NegEx-annotated phrase (Kocbek et al. 2016)**

### 4.2.2 Non-Ontology-Based Approaches

Some studies proposed an alternative to ontology based IE. Rochefort et al. (2015) proposed a document-by-term matrix, which transforms text into a structured format. The basic idea is that each document becomes a term vector where each term is an attribute of the vector. The value of each component equals the number of times the corresponding term occurs in the document. More recent studies have focused on „Bag-Of-Words' (BOW), which represents a multiset of words (unigram) regardless of word order or grammar (Agarwal et al. 2016, Kocbek et al. 2016). Unigrams ($n = 1$) simplify the problem, but poses a major shortcoming as semantics of the words get lost. To add context and retain local dependencies, several studies introduced higher order n-grams, which describe a sequence of $n$ consecutive items from a given sequence of medical text (Rochefort et al. 2015). Expanding the bag concept to 'Bag-Of-Phrases' (BOP) is another way to maintain contextual knowledge (Kocbek et al. 2016). This is in line with Martinez et al. (2015) who applied the JulieLab tool to segment text into sentences. They found that model performance based on BOP features outperformed the one based on BOW.

Tagging phrases according to their data source may be important when multiple reports address different disease separately. When a patient with lung problems undergoes radiology and pathology tests, to tested for pneumonia and breast cancer, the radiology report contains the words 'lung', 'pain', and 'cough' while the pathology report contains 'cancer', 'breast' and 'tumor'. This poses a major risk of misclassifying patients, e.g. with 'breast cancer'. Tagging phrases according to their source also allows to identify sources that contain the most valuable information.

Elimination of punctuation and the conversion to lower case was proposed by Rochefort et al. (2015) while Martinez et. al (2015) marked the position of questions marks (end, inside, or beginning of a sentence) to identify speculative sentences. Some studies used tokenization to convert a sequence of character into single pieces (Kasthurirathne et al. 2016, Martinez et al. 2015). Additionally, the application of Lingua module helps to identify stop words like 'and', 'the' or 'maybe'. Stopwords that occur in the final token list could be removed to improve predictive performance as they do not contain any useful information (Kasthurirathne et al. 2016). One advantage of tokenization is that it allows for the identification and removal of tokens with low prevalence. However, removal of stop words may create potential shortcomings when documents are searched for an exact sentence or phrase rather than a set of keywords.

## 4.3 Data Reduction

Data reduction is used to improve predictive performance by reducing the number of features to the most relevant for the predictive task. It serves a variety of purposes:

- Prevent dimensionality
- Reduce training time of the model
- Potential to increase interpretability

- Reduce overfitting

Extensive research has focused on rather two different ways to identify features that will classify a cohort with a certain medical condition: (1) manual methods which rely on expert knowledge and (2) statistical (machine learning) methods to select and optimize a subset of features which will most accurately predict a phenotype cohort.

### 4.3.1 Manual Feature Selection

Manual feature selection refers to the application of expert algorithms. An expert is someone with extensive medical knowledge (e.g. physician) who defines logical constraints (rules) based on their expertise. Known risk factors can often indicate the presence of a certain phenotype. Those include covariate (e.g. sex, age, smoking status), laboratory tests and vital signs (Evans et al. 2016, Wang et. al 2016, Anderson et al. 2015). They typically define a threshold (e.g. blood glucose levels) as proxy (Zhong et al. 2016) while others use medications (Spratt et al. 2016) to classifying patients with a certain medical condition. Another common technique is to select disease related ICD-9 codes (Zhong et al. 2016, Osbourne et al. 2016, Teixeira et al. 2016).

When dealing with unstructured data, manual feature selection only plays a minor role. Kasthurirathne et al. (2016) asked two clinicians to manually select and rank tokens that would indicate the presence of cancer. They compared results and in case of disagreement resolved the issue by consulting a third expert. Wei et al. (2015) followed the same approach, except their multi-class classifier required a total of five medical experts to review all features. Features were reviewed and selected by at least one medical expert. Cohens Kappa (*k*) was used to measure inter-rater agreement, which is more robust than a simple percent agreement, since *k* takes into consideration the probability of the agreement occurring by chance. Some studied searched for specific keyword, e.g. 'cancer', to identify patients with a certain medical condition (Osbourne et al. 2016).

### 4.3.2 Algorithmic Feature Selection

Algorithmic feature selection is not a formally defined term but for the purpose of the paper will refer to an algorithmic process of selecting relevant features from the original dataset. Feature selection is a two-step process that generally involves feature ranking and subset selection.

To obtain specific disease related medications Wei et al. (2015) used medications defined by the MEDication Indication (MEDI), which lists medications and their estimated prevalence. They selected medication for each disease with prevalence $\geq 80$ percent, which ensure the selection of highly specific medications and thus increases accuracy. Padula et al. (2016) proposed a regression of covariates of to identify and select for pressure ulcer. Evans et al. (2016) proposed a descriptive statistical analysis to identify predictors for heart failure based on patients who had a primary diagnosis of heart failure.

Some studies listed terms by discriminating power as measured by information gain (IG), also referred to Kullback-Leibler divergence, which measures the effect of a terms on the entropy of the target label (Kasthurirathne et al. 2016, Kocbek et al. 2016). A common technique is to use token with the n-highest IG as features to identify patients with the predefined medical condition. Kocbek et al. (2016) selected the top 10 features and gradually increased the number to the full set of 50 features. They found that decreasing selected features may actually improve predictive performance.

Another technique, which was used by Agarwal et al. (2016), is to list terms by frequency and selected the ones accounting for more 90% of all patients. The selected terms were reviewed to remove ambiguous terms. For example, synonyms for ‚T2DM' (Type Two Diabetes Mellitus) include terms like ‚MODY' (mature onset diabetes of the young). However, current medical practice makes a distinction between T2DM and MODY. Therefore, removing MODY from the list is necessary to improve specificity of the predictive model.

Rochefort et al. (2015) proposed a 10-fold cross-validation approach to select a feature subset from the original 62.416 distinct features. They performed feature selection within each of the k-1 training folds and used Person's correlation coefficient to identify features associated with the predefined medical condition (in this case thrombosis). The final list of the top 30 features was then selected for the predictive model.

### 4.3.3 Algorithmic Feature Extraction

Algorithmic feature extraction is not a formally defined term but for the purpose of the paper will refer to an algorithmic process of building a new set of features from the original dataset. Feature extraction aims to reduce dimensionality by combining features instead of deleting them. The result is a set of fewer features with new values.

Sources with the same type of features (e.g. ‚outpatient diagnosis reports' and ‚inpatient diagnosis records') may contain the same information. Hence, they are highly correlated and therefore will negatively impact the classification model. Zeng et al. (2017) addressed this issue by summarizing sources that were highly correlated and merging similar features within those sources. This process reduced the number of features from 36 to 8 and increased the predictive performance of the model. Choi et al. (2016) proposed a similar approach by grouping diagnosis codes and medications codes into once feature vector. However, feature extraction has played a minor role only.

## 5  Model selection

|  | Features | | | Classification algorithm | | | | | | Number of classes | | | |
|---|---|---|---|---|---|---|---|---|---|---|---|---|---|
|  |  | Algorithmic | | | | | | | | | Multiclass | | |
|  | Manual | Selection | Extraction | Ensemble | Tree-based | SVM | NLP | Log. Reg. | Other | 1 | 2 | 5 | 9 |
| **Diabetes** | 3 | 2 | 2 | 2 | 3 | 1 | 1 | 2 | 1 | 4 | 1 | - | 0,1 |
| **Cancer** | 2 | 3 | 3 | 2 | 1 | 2 | 2 | 1 | - | 5 | - | 0,2 | 0,1 |
| **Heart failure** | 1 | - | 3 | 1 | - | - | - | 2 | 1 | 3 | 0,5 | - | - |
| **Atrial Fibrillation** | 1 | 2 | - | - | 1 | - | 1 | - | 1 | 1 | 0,5 | - | 0,1 |
| **Arthritis** | - | 1 | 1 | - | - | - | 1 | 1 | - | - | 0,5 | - | 0,1 |
| **Thrombosis** | 1 | - | 1 | - | - | 2 | - | - | - | 1 | - | 0,2 | - |
| **Others** | 1 | 2 | 4 | 4 | 1 | 2 | 2 | 2 | - | 4 | 0,5 | 0,6 | 0,5 |
| ∑ | 9 | 10 | 14 | 9 | 6 | 7 | 7 | 8 | 3 | 18 | 3 | 1 | 1 |

**Table 5** Data mining approaches

*Ensemble* is a widely used method which uses several learning algorithms to obtain a better predictive outcome. Wang et al. (2015) trained and evaluated random forest classification models to detect heart failure. Although random forest is not necessarily the best classification method, it is good at handling nonlinear interactions. The combination of structured and unstructured data performed superior to using structured or unstructured data alone. Teixeira et al. (2016) also proposed a random forest model which used a variety of structured and unstructured data. They found random forests performed better at classifying subjects with hypertension than deterministic algorithms. They also demonstrated all categories of identification performed superior to blood pressure measurement, which is the standard procedure for detecting hypertensive individuals. This might be due to treatment reducing blood pressure to normal level or other causes of non-hypertension related high blood pressure which can be found in EHR. Padula et al. (2016) developed a method to classify patients with pressure ulcers (also decubitus). They proposed random forests to reduce high-dimensionality and a multivariate logistic regression to evaluate the associations between covariates (e.g. medications, lab tests) and the phenotype of interest. Anderson et al. (2015) used diagnosis, medications and clinical notes to identify patients with type two diabetes. They used a combination of random-forests probabilistic models and multivariate logistic regression which both performed better than the conventional model mimicking conventional risk scores. They also tested incomplete EHR by excluding medications and observed almost no changes in predictive performance. This suggests diagnosis contain similar information as medications prescribed to treat them. Kinar et al. (2016) used basic demographic data (age, sex) and blood counts to detect cancer cases. They combined a gradient boosting model with a random forest model and found the best performing algorithm was an ensemble of decision trees.

*Support Vector Machine (SVM)* is machine learning algorithm, which can be applied for both linear and non-linear tasks (Hastie et al. 2009). SVM have shown high performance in classifying disease such as deep vein thrombosis, pulmonary embolism (Rochefort et al. 2015) and various types of cancer (Kocbek et al. 2015, Kocbek et al. 2016). The basic idea is to construct a high dimensional feature space by applying kernel functions, which replaced every data point with a nonlinear function. The model then looks for optimal parameters which control the trade-off between false negatives and false positives, as well as the shape of the hyperplane. Rochefort et al. (2015) proposed 10 alternative SVM to identify deep vein thrombosis and pulmonary embolism from radiology reports. They The model then looked for optimal parameters which controlled the trade-off between false negatives and false positives, as well as the shape of the hyperplane. Kocbek et al. (2015) proposed a SVM which and showed that a combination of structured and unstructured data would improve model performance. Based on those findings Kocbek et al. (2016) proposed a method which further investigates the effect of linking multiple data sources on text classification. They proposed a SVM to solve a multi-class problem which confirmed previous findings: data linkage leads to an improvement in predictive performance.

*Natural Language Processing (NLP)* was proposed by a number of studies. Wei et al. (2015) used a combination of ICD-9 codes, medications and narrative text (e.g. admission notes, discharge summaries) to identify a variety of diseases (e.g. atrial fibrillation, Alzheimer's disease, breast cancer, gout, HIV, multiple sclerosis). Clinical notes were searched for disease specific keywords, while negations (e.g. 'no cancer') were excluded. Medications were defined using MEDication Indication (MEDI), which lists disease related medications and their estimated prevalence. To ensure high specificity only medications with prevalence ≥ 80% were selected. Additionally, the MedEx NLP system was used to extract medication names and other signature from clinical notes. The showed the combination of ICD-9 codes, medications and narrative text can improve predictive performance, thus the use of ICD-9 codes alone may not be sufficient. Osbourne et al. (2016) proposed a Cancer Registry Control Panel (CRCP) which used standard NLP infrastructure 'based on Unstructured Information Management Architecture, […] North American Association of Central Cancer Registries (NAACCR) search criteria, and Facility Ontology Registry Data Standards terms' to identify cancer cases. The model extracted concepts and identify mentions of cancer in pathology reports. They found that NLP is sufficient to identify patients with cancer, although additional use of ICD-9 codes may improve predictive performance.

*Tree-based* approaches another type of supervised learning. Yahi & Tatonetti (2015) proposed an 'Ontology-driven Reports-based Phenotyping from Unique Signatures' (ORPheUS) to detect both, type two diabetes and atrial fibrillation. Their model used abnormal lab tests from clinical pathology reports and scanned for matches in patient histories. Zhong et al. (2016) used a combination of lab tests, medications and ICD-9 codes to identify patients with diabetes type one and type two. They proposed a Classification and Regression Tree (CART) algorithm which has no underlying assumption about variable distribution or relationship. Rules were manually defined and based on expert knowledge (e.g. blood glucose ≥ 200). They demonstrated usage of ICD billing codes alone was the best way to detect type one diabetes and diabetes in general. Classifying type two diabetes, using ICD-9 codes alone, may not be possible since ICD-9 type two codes includes 'unspecified' diabetes.

*Logistic Regression* was used by some studies. Statistical models can yield high recall and precision, but require expert assignment of phenotype labels to patient records (Peissig et al. 2014). Logistic Regressions does not rely on the assumption of normal distribution and is therefore easy to implement. Evans et al. (2016) proposed a multivariate logistic regression model to classify high-risk heart failure patients. Features were extracted from a variety of data, entered into the model and removed stepwise using a backward elimination. High-dimensionality is a common problem in EHR. A divergence between the number of variables and the number of observations often leads to the problem of collinearity[5]. To solve this issue Agarwal et al. (2016) trained and tested a L1 penalized logistic regression model, referred to as EXPRESS (eXtraction of Phenotypes from Records using Silver Standards). They hypothesized a large amount of labeled data can offset an inaccuracy in the labels and demonstrated feasibility of detecting patients at risk for heart failure using noisy labeled training data. Yu et al. (2015) proposed an adaptive elastic-net penalized logistic regression model, which uses L1 und L2 penalization to help balance the 'in-sample predictive accuracy and the model complexity'. The model used an online knowledge and a variety of EHR data to identify individuals with type two diabetes and myocardial infarction.

---

[5] occurs when the number of parameters exceeds the number of observations

*Comparison of popular algorithms* was done in several studies, which developed different classification models and compared their performance. Kasthurirathne et al. (2015) proposed several ML models, such as logistic regression, naïve bayes, k-nearest neighbor, random forest and decision tree, to identify patients with cancer. Performance of k-nearest neighbor and naïve bayes significantly declined by adding more features (20 as oppose to 5, 10 or 15 features) to the model. This might be due to the underlying assumption of independence of features which might not always be given. This is in line with Zheng et al. (2017) who demonstrated an improvement in predictive performance when features are summarized into higher levels. They added a SVM model to the algorithms proposed by Kasthurirathne et al. (2015) and compared the model's ability to classify individuals with type two diabetes. Performance of random forest, decision tree and SVM was more stable compared to the remaining three classifiers. Martinez et al. (2015) compared SVM, Bayesian Network models, Naive Bayes and Random Forest while SVM performed best at classifying invasive fungal diseases. Ichikawa et al. (2016) proposed several ML methods to identify patients with hyperuricemia, which is a surplus of uric acid in the blood and in turn used to diagnose gout. They compared gradient-boosting decision tree, random forest and logistic regression models, and demonstrated that all ML methods showed similar predictive performance.

# 6   Conclusions

The paper examined and addressed the following three research questions:

*1. "What data generate the best predictive outcome?"*

A development can be seen from single to multi data-source usage. Previous research has focused on single data sources to identify patient cohorts (Hripcsak et al. 2002, Nguyen et al. 2010). Most of these models relied on ICD-9 codes which would indicate the presence or absence of a certain medical condition. However, there are a multitude of potential shortcoming to using ICD-9 codes as they lack specificity in the information transmitted and can thus lead to misclassification or even missing patients. Therefore, a number of studies have investigated the use of additional data sources with a strong emphasis on narrative text, which is usually hidden in the free text. Hence, extracting information requires additional tools such as NLP. Progress of medical NLP is slower compared to NLP in other fields. This is due to challenges involved in the processing of clinical notes such as poor access to shared data, lack of annotated datasets, lack of standards for annotation and limited collaboration among others (Aggarwal & Reddy 2015, p 238). However, combining structured and unstructured data has been shown to outperform using structured or unstructured data alone (Wang et al. 2015). Therefore, additional use of ICD-9 codes could improve classification problems significantly (Osborne et al. 2016). Nonetheless, how these sources are assembled and applied affects how many patients can be detected significantly (Spratt et al. 2016).

*2. Which machine learning approaches achieve the best results?*

In general, research suggests that ML is a more accurate and efficient approach to the detection of phenotypes, than expert review (Zeng et al. 2017, Ichiwaka et al. 2016). However, comparing model performance directly is challenging due to the diversity of population, in terms of origin, timeframe and sizes. Benchmark studies compared different ML approaches based on the same dataset. Performance of Naïve bayes and KNN tended to show a statistically significant decline in performance when feature size was increased (Kasthurirathne et al. 2016). However, predictive performance of random forest and simple logistic regression improved when features were summarized into higher levels (Kasthurirathne et al. 2016). This is in line with Zheng et al. (2017) who showed similar results for random forest approaches. It also indicates that increasing levels of feature abstraction may lead to an increase in model performance. Furthermore, decision tree and SVM were more stable compared to artificial neural nets and logistic regression (Zheng et al. 2017). In some cases, SVM performed best compared to Bayesian Network models, Naive Bayes and Random Forest (Martinez et al. 2015). In other cases, deep learning models have shown to exceed the performance of common models such as logistic regression and K-nearest neighbor (Choi et al. 2016). The performance of automated feature extraction exceeded expert review and may be a way towards high-throughput phenotyping (Yu et al. 2015). Yet, the variety of data makes it difficult to compare performance of ML algorithms.

*3. What are potential gaps for further research?*

NLP of clinical text is a growing trend in the area of ML. Yet, progress of NLP in the medical area is slower compared to NLP in general. This is due to some challenges that present themselves in the processing of clinical notes. These include poor access to shared data, lack of annotated datasets, lack of standards for annotation and limited collaboration among others (Aggarwal & Reddy 2015, p 238). Future research should focus on addressing some of these issues by embracing collaboration and developing NLP standards. A common database could help to improve comparability to further investigate which ML modules is best suited to which type of medical data.

# 7 Appendix

| | Features | | | Classification algorithm | | | | | | Number of classes[6] | | | |
|---|---|---|---|---|---|---|---|---|---|---|---|---|---|
| | | Algorithmic | | | | | | | | | Multiclass | | |
| | Manual | Selection | Extraction | Ensemble | Tree-based | SVM | NLP | Log. Reg. | Other | 1 | 2 | 5 | 9 |
| **Diabetes** | | | | | | | | | | | | | |
| Agarwal et al. (2016) | | | 1 | | | | | 1 | | | 0,5 | | |
| Anderson et al. (2015) | 1 | | | 1 | | | | | | 1 | | | |
| Spratt et al. (2016) | 1 | | | | | | | | 1 | 1 | | | |
| Wei et al. (2015) | 1 | | | | | | 1 | | | | | | 0,1 |
| Yahi & Tatonetti (2015) | | 1 | | | 1 | | | | | | 0,5 | | |
| Zeng et al. (2017) | | | 1 | 1 | 1 | 1 | | 1 | | 1 | | | |
| Zhong et al. (2016) | 1 | | | | 1 | | | | | 1 | | | |
| ∑ | **4** | **1** | **2** | **2** | **3** | **1** | **1** | **2** | **1** | **4** | **1** | **-** | **0,1** |
| **Cancer** | | | | | | | | | | | | | |
| Kasthurirathne et al.2016 | 1 | 1 | | 1 | 1 | | | 1 | | 1 | | | |
| Kinar et al. (2016) | | 1 | | 1 | | | | | | 1 | | | |
| Kocbek et al. (2015) | | | 1 | | | 1 | | | | 1 | | | |
| Kocbek et al. (2016) | | | 1 | | | 1 | | | | | | 0,2 | |
| Li et al. (2015) | | | 1 | | | | | | | 1 | | | |
| Osborne et al. (2016) | 1 | | | | | | 1 | | | 1 | | | |
| Wei et al. (2015) | | 1 | | | | | 1 | | | | | | 0,1 |
| ∑ | **2** | **3** | **3** | **2** | **1** | **2** | **2** | **1** | **-** | **5** | **-** | **0,2** | **0,1** |
| **Heart failure** | | | | | | | | | | | | | |
| Agarwal et al. (2016) | | | 1 | | | | | 1 | | | 0,5 | | |
| Choi et al. (2016) | | | 1 | | | | | | 1 | 1 | | | |
| Evans et al. (2016) | 1 | 1 | | | | | | 1 | | 1 | | | |
| Wang et al. (2015) | | | 1 | 1 | | | | | | 1 | | | |
| ∑ | **1** | **1** | **3** | **1** | **-** | **-** | **-** | **2** | **1** | **3** | **0,5** | **-** | **-** |

---

[6] $class\ weight = \frac{1}{[total\ no.\ of\ classes] * [no.\ of\ considered\ classes]}$

|  | Features | | | Classification algorithm | | | | | | Number of classes | | | |
|---|---|---|---|---|---|---|---|---|---|---|---|---|---|
|  | | Algorithmic | | | | | | | | | Multiclass | | |
|  | Manual | Selection | Extraction | Ensemble | Tree-based | SVM | NLP | Log. Reg. | Other | 1 | 2 | 5 | 9 |
| **Atrial Fibrillation** | | | | | | | | | | | | | |
| Yahi & Tatonetti (2015) |  | 1 |  |  | 1 |  |  |  |  |  | 0,5 |  |  |
| Wang et al. (2016) | 1 |  |  |  |  |  |  |  | 1 | 1 |  |  |  |
| Wei et al. (2015) |  | 1 |  |  |  |  | 1 |  |  |  |  |  | 0,1 |
| ∑ | **1** | **2** | **-** | **-** | **1** | **-** | **1** | **-** | **1** | **1** | **0,5** | **-** | **0,1** |
| **Arthritis** | | | | | | | | | | | | | |
| Wei et al. (2015) |  | 1 |  |  |  |  | 1 |  |  |  |  |  | 0,1 |
| Yu et al. (2015) |  |  | 1 |  |  |  | 1 |  |  |  | 0,5 |  |  |
| ∑ | **-** | **1** | **1** | **-** | **-** | **-** | **1** | **1** | **-** | **-** | **0,5** | **-** | **0,1** |
| **Thrombosis** | | | | | | | | | | | | | |
| Kocbek et al. (2016) |  |  | 1 |  |  | 1 |  |  |  |  |  | 0,2 |  |
| Rochefort et al. (2015) |  | 1 |  |  |  | 1 |  |  |  | 1 |  |  |  |
| ∑ | **-** | **1** | **1** | **-** | **-** | **2** | **-** | **-** | **-** | **1** | **-** | **0,2** | **-** |
| **Others** | | | | | | | | | | | | | |
| Ichikawa et al. (2016)[7] | 1 |  |  | 1 |  |  |  | 1 |  | 1 |  |  |  |
| Kocbek et al. (2016)[8] |  |  | 1 |  |  | 1 |  |  |  |  |  | 0,6 |  |
| Martinez et al. (2015)[9] |  |  | 1 | 1 | 1 | 1 |  |  |  | 1 |  |  |  |
| Padula et al. (2016)[10] |  | 1 |  | 1 |  |  |  |  |  | 1 |  |  |  |
| Teixeira et al. (2016)[11] |  |  | 1 | 1 |  |  | 1 |  |  | 1 |  |  |  |
| Wei et al. (2015)[12] |  | 1 |  |  |  |  | 1 |  |  |  |  |  | 0,5 |
| Yu et al. (2015)[13] |  |  | 1 |  |  |  | 1 |  |  | 1 | 0,5 |  |  |
| ∑ | **1** | **2** | **4** | **4** | **1** | **2** | **2** | **2** | **-** | **4** | **0,5** | **0,6** | **0,5** |

---

[7] Classifies hyperuricemia

[8] Classifies cancer, thrombosis, secondary malignant neoplasm of respiratory and digestive organs, multiple myeloma and malignant plasma cell neoplasms and pneumonia

[9] Classifies fungal disease

[10] Classifies decubitus

[11] Classifies hypertension

[12] Classifies diabetes, cancer, atrial fibrillation, Arthritis, Alzheimer, gout, HIV, MS and Parkinson

[13] Classifies arthritis and coronary artery disease

| author | journal | place, time | baseline | validation | data source | conclusion |
| --- | --- | --- | --- | --- | --- | --- |
| Agarwal et al. (2016) | JAMIA | USA, 1994-2013 | expert review | training, testing | academic medical center | noisy training data for a learning models may be a feasible alternative to manually labeled training data |
| Anderson et al. (2015) | JBI | USA, 2009-2012 | calculated ground truth | cross-validation | health care system | strong predictive power of EHR when excluding medications which suggests that diagnosis contain the same information as medications |
| Choi et al. (2016) | JAMIA | USA, 2010-2013 | calculated ground truth | training, testing | academic medical center | deep learning model considering the sequential nature of EHR exceed the performance of common models such as logistic regression and K-nearest neighbor |
| Evans et al. (2016) | JAMIA | USA, 2014 | expert review |  | health care system | computer decision support system can improve patient heart failure identification and decrease mortality |
| Ichikawa et al. (2016) | JBI | Japan, 2011-2013 | calculated ground truth | training, testing | health insurance | Suggest that all ML methods have similar capability in predicting hyperuricemia and under sampling did not significantly improve predictability |
| Kasthurirathne et al. (2016) | JBI | USA | expert review | training, testing | health care system | suggests an alternative for automated cancer detection form free text pathology reports, using non-dictionary modeling approaches |
| Kinar et al. (2016) | JAMIA | UK, 1990-2012 Israel, 2003-2011 | reference cohort | training, testing | health care system | demonstrate early detection of cancer by analyzing sex, age and blood count |
| Kocbek et al. (2015) | An. Con. in Big Data and Health | Australia, 2012-2014 | ICD-10 codes | cross-validation | health care system | combination of multiple data sources improves performance of lung cancer classification |
| Kocbek et al. (2016) | JBI | Australia, 2012-2014 | expert review | cross-validation | health care system | combining data sources improves predictability, but the best combination of data sources depends on the specific phenotype to be identified |
| Li et al. (2015) | JBI | USA | calculated ground truth | cross-validation | academic medical center | suggest that bag-level features, which describe the characteristics of the entire image, and instance-level features which denote the texture and intensity, equally contribute to the final classification of gastric cancer |
| Martinez et al. (2015) | JBI | Australia, 2003-2011 | expert review | cross-validation | academic medical center | ML offers the opportunity for real-time surveillance of fungus infections using free-text from CT scans |
| Osborne et al. (2016) | JAMIA | USA, 2014-2015 | expert review | training, testing | government agency | NLP is sufficient to identify patients with cancer, although additional use of ML and ICD-9 codes may improve performance |

| author | journal | place, time | baseline | validation | data source | conclusion |
| --- | --- | --- | --- | --- | --- | --- |
| Padula et al. (2016) | JAMIA | USA, 2011-2014 | ICD-9 codes | cross-validation | academic medical center | indicates feasibility of early detection of pressure ulcers |
| Rochefort et al. (2015) | JAMIA | Canada, 2008-2012 | expert review | cross-validation | academic medical center | NLP can be used to detect thrombosis from radiology reports with high accuracy |
| Spratt et al. (2016) | JAMIA | USA, 2007-2011 | expert review | training, testing | academic medical center | phenotypes can be detected from various sources, but how these sources are assembled and applied effects how many patients can be detected |
| Teixeira et al. (2016) | JAMIA | USA, 2007-2009 | expert review | training, testing | academic medical center | combination of multiple EHR data significantly improves predictability for hypertension (as oppose to ICD-9 codes only) |
| Wang et al. (2015) | Conf Proc IEEE Eng Med Biol Soc. 2015 | USA, 2001-2010 | ICD-9 codes | cross-validation | health care system | combination of structured and unstructured data can facilitate early detection of heart failure and is superior to using structured or unstructured data alone |
| Wang et al. (2016) | JAMIA | USA, 2009-2011 | expert review | cross-validation | academic medical center | rule based algorithms can be used to identify individuals with chronic disease via known risk factors and help facilitate targeted interference |
| Wei et al. (2015) | JAMIA | USA | expert review | | academic medical center | demonstrates that multiple EHR sources improve predictive performance, thus the use of ICD-9 codes may not be sufficient |
| Yahi, Tatonetti (2015) | PAMIA | USA | eMERGE | cross-validation | academic medical center | pathology reports may be an effective screening tool for automated phenotype detection |
| Yu et al. (2015) | JAMIA | 2014 | expert review | training, testing | online knowledge | performance of automated feature extraction exceeds expert review and may be a way toward high-throughput phenotyping |
| Zeng et al. (2017) | IJMA | China, 2012-2014 | expert review | cross-validation | government agency | suggests that ML is a more accurate and efficient approach to the detection of phenotypes, then expert review (it also indicates increasing level of feature abstraction may lead to an increase in model performance) |
| Zhong et al. (2016) | JAMIA | USA, 2011-2012 | expert review | cross-validation | academic medical center | EHR may be used for large scale surveillance of childhood diabetes, requiring minimal effort of manual review |